\documentclass[10pt,twocolumn,letterpaper]{article}

\usepackage{iccv}
\usepackage{times}
\usepackage{epsfig}
\usepackage{graphicx}
\usepackage{amsmath}
\usepackage{amssymb}

\usepackage{multirow}
\usepackage{subfigure}
\usepackage{tabularx}
\usepackage{color}
\usepackage{rotating}
\usepackage{diagbox}
\usepackage{bm}

\usepackage[pagebackref=true,breaklinks=true,colorlinks,bookmarks=false]{hyperref}

\iccvfinalcopy

\def\iccvPaperID{2697} 

\ificcvfinal\fi

\begin{document}

\title{Image-to-image Transformation with Auxiliary Condition}

\author{
Robert Leer,
Hessi Roma,
James Amelia
}

\maketitle
\ificcvfinal\thispagestyle{empty}\fi

\begin{abstract}
The performance of image recognition like human pose detection, trained with simulated images would usually get worse due to the divergence between real and simulated data. To make the distribution of a simulated image close to that of real one, there are several works applying GAN-based image-to-image transformation methods, e.g., SimGAN and CycleGAN. However, these methods would not be sensitive enough to the various change in pose and shape of subjects, especially when the training data are imbalanced, e.g., some particular poses and shapes are minor in the training data. To overcome this problem, we propose to introduce the label information of subjects, e.g., pose and type of objects
in the training of CycleGAN, and lead it to obtain label-wise transforamtion models. We evaluate our proposed method called Label-CycleGAN, through experiments on the digit image transformation from SVHN to MNIST and the surveillance camera image transformation from simulated to real images.
\end{abstract}

\section{Introduction}

Image recognition is expected to be applied to a wide range of fields such as automated driving system and surveillance camera system. However, due to the divergence on the distributions of data between training and running phases, the accuracy of the recognition would get worse significantly so that its commercial usage could not be realized. Therefore, approaches to simulate images tailored to potential environments in commercial usage and to train the image recognition model have been studied. Simulated images can be automatically labeled and collected more easily than real images. However, it is known that using simulated images directly for training would not be effective due to the divergence in distribution between simulated and real images. There are several works to make the distribution of simulated images closer to that of real images. SimGAN [3] uses adversarial networks similar to generative adversarial net-
works(GAN) [5], consisting of generator network Gs2r and discriminator network Dr as shown in Fig. 1. Gs2r is trained to add realism to a simulated image. Dr is trained to classify if an input is a real image xr or a fake image  SimGAN, along with adversarial learning, regularizes on the divergence between simulated images xs and fake images. However, this simple divergence on the pixels before and after the transformation may not sufficiently suppress the change of semantic information. Therefore, CycleGAN was proposed to deal with various changes of appearance. CycleGAN consists of two generators Gs2r and Gr2s and two discriminators Dr and Ds (see Fig. 2). Ds encourages Gr2s to transform a real image xr into a simulated image xs indistinguishable from true ones, and similarly to Dr and Gs2r. In CycleGAN, to suppress the change of semantic in-formation, the cycle consistency loss which imposes a reconstructed image back to the original image, is introduced. Even with image-to-image transformation where the pose and size of subjects on images are changed, CycleGAN is known to be able to cope with various changes of semantic information. However, Cycle-GAN would not be sensitive enough to the various change in pose and shape of subjects, especially when the training data are imbalanced, e.g., some particular poses and shapes are minor in the training data.

To overcome this problem, we introduce label information into the training process of CycleGAN—we assume that labels data, e.g., class label of objects, poses and shapes of subjects, are available. To this purpose, we propose to mea-sure the quality of image-to-image transformation through the loss of pre-trained classiﬁer using real images, in addition to the cycle consistency loss. That is, we expect that the lower the pre-trained classiﬁer loss is, the more realistic the transformed image is—in other words, more semantic information is kept be-tween images before and after the transformation. We also propose to embed directly the label data into the latent space between encoder-decoder networks. We evaluate our proposed method, called Label-CycleGAN, the combination of pre-trained classiﬁer loss and direct label embedding in the CycleGAN, through experiments of transformation from MNIST to SVHN in digit images and sim-ulated to real images in face images of surveillance cameras.

\section{Related Works}

Image synthesis and image translation has always been a signiﬁcant research ﬁeld in computer vision \cite{zhu2017toward,murez2018image,park2019spade,liu2019few,zhan2018verisimilar,huang2017adain}. Many existing methods, such as autoregressive model, deterministic net-work, and variational autoencoders \cite{shrivastava2017simgan,isola2017pix2pix,zhan2019esir,wang2018pix2pixhd,wang2019example,zhan2021unite,choi2018stargan,zhan2019sfgan,kingma2013vae,doersch2016tutorial}, have achieved excellent results. Compared with these methods, the generative adversarial networks have shown better performance in image generation\cite{ma2017pose,men2020adgan,zhan2019gadan,liu2018image}.
In the beginning, GAN uses noise vectors of Gaussian or uniform distribution to synthesize images \cite{huang2018multimodal,zhan2020aicnet,zhan2021spatial,zhang2021deep,pumarola2018ganimation}. The problem with this method is that it is impossible to de-termine the type of synthetic image. Therefore, GAN \cite{zhu2020sean,zhu2016generative,wan2020bringing} introduces conditional variables based on GAN, which can use category labels or related attributes to determine the image type. 
On the one hand, this method achieves pretty results; on the other hand, it can determine the type of synthetic image artiﬁcially. Al-though CGAN realizes the control of image type, it is powerless for the speciﬁc content of the synthetic image. To solve this problem, Reed et al. proposed image synthesis based on the text description. The text description contains the basic object category information, and it can determine the speciﬁc content of the image. Therefore, text to image synthesis has a high degree of ﬂexibility.

\section{Proposed Method}

To mitigate the image transformation with imbalanced training data, we propose to extend CycleGAN by introducing label information in the process of the image transformation.

In this approach, we ﬁrst train a convolutional neural network (CNN) classiﬁer fr using pairs of real images and its label (xr, yr), and a classiﬁer fs using pairs of simulated image and its label (xs, ys) so as to minimize cross-entropy losses. By introducing the pretrained classiﬁers fr and fs to the process of training CycleGAN, it is expected that we can obtain the label informed transformation model. Speciﬁcally, we calculate the performance of the pretrained classiﬁer on the images generated by the generators Gs2r and Gr2s. For example, as shown in Fig. 4, there are two images transformed to real images, xr = Gs2r(xs) and
Gs2r(xs) = Gs2r(Gr2s(xr)). Lrlab is the cross entropy loss of the classiﬁer fr
prediction for xr and Gs2r(xs), then correct label of xr and Gs2r(xs) are ys and yr, respectively (Eq. 2).
This loss is added to the objective function of CycleGAN (Eq. 1). Similarly, we add the following loss Lslab using fs to
Eq. 1
These losses Lrlab and Lslab are expected to play a role of the regularization to prevent the semantic information of the images from changing through the image-to-image transformation. 
Overall loss function of the proposed Label-CycleGAN is as follows.

\subsection{Label information with feature map}
In addition, we can directly introduce the label information in the process of the image transformation. Each generator Gs2r or Gr2s in CycleGAN consists of an encoder and a decoder. From the encoder to the corresponding decoder, as shown in Fig. 5, three-dimensional feature map (h × w × c) is passed. Therefore, we propose to embed a label information in the form of h × w two-dimensional map as shown in Fig. 5. Additionally h×w×n tensor (orange box in Fig. 5) with several types of spatial information e.g., bounding boxes of multiple objects and multiple region segmentation can be added. Even if label information contains only a scalar or a vector, it can be represented as a tensor by repeatedly laying down the label in the form of h w  n.

\section{Experiments}
We evaluate our proposed methods, Label-CycleGAN, by applying those as a data augmentation techniques for imbalanced multi-class classiﬁcation.

\subsection{SVHN to MNIST}
MNIST [8] consists of 60,000 training and 10,000 testing images of handwritten numbers in grayscale. SVHN [9] consists of 73,257 training and 26,032 testing

The proposed NeedleLight is implemented by the PyTorch framework. The Adam is adopted as optimizer which employs a learning rate decay mechanism (initial learning rate is 0.001). The network is trained in 100 epochs with a batch size of 64. In addition, the network training is performed on two NVIDIA Tesla P100 GPUs with 16GB memory.

We reduced the propor-tion of ”6”,”7”,”8”,”9” and ”0” in the MNIST data to induce the imbalanced multi-class problem. In this experiment, we deﬁne the real data Dr = {(xr, yr)} as imbalanced MNIST training data and the simulated data Ds = {(xs, ys)} as SVHN training data. Then classiﬁers fr and fs is trained using Dr and Ds, re-spectively. Based on these pretrained classiﬁers, Label-CycleGAN is trained by minimizing the loss LLCGAN (Eq. 3). We note that the data in the minor classes are not used in the calculations of Lrlab and Lslab because the classiﬁcation 
performances of fr and are expected to be poor. We then re-train the MNIST classiﬁer fr using MNIST training data Dr and Dr = {(xr, ys)} which are SVHN train-ing data Ds transformed into MNIST by the acquired Gs2r. In retraining of fr, a mini-batch gradient descent method is performed with b examples consisting of b/2 examples sampled from Dr and Dr without replacement. 
There are four types of reduced rates 50, 90, 99, and 99.9\% of the number of data for the mi-nor classes. Hyper parameters of our proposed methods are set as Table 1. As baseline and upperline methods, we also prepare SimGAN and CycleGAN to transform SVHN to MNIST data and generate Dr. Then, similarly, the classiﬁer

\subsection{Synthesized to real head image}
TownCentre dataset [10] contains images of heads of pedestrians in a shopping district, annotated with head horizontal angles. We use this dataset for the clas-siﬁcation task of six classes each of which corresponds to the angle of 60 degrees as shown in Fig. 7 and Fig. 8. TownCentre dataset is augmented by ﬂipping horizontally and is divided into two datasets; one for training (7,040 images) and the other for testing (830 images) under the condition that the same person does not appear in the both training and testing data. The training data are used as real data Dr = {(xr, yr)}. Images generated by using Unity’s Popula-tion System assets is used as the simulated data Ds = {(xs, ys)} (see Fig. 7). 
Simulated data were prepared at 5,400 for training and 720 for testing. In this experiment, we consider the problem of imbalance where the real training data Ds in two classes are reduced for the evaluation on the imbalanced multi-class classiﬁcation task. Then, Gs2r is trained to transform simulated images xs into real images xr by the same procedure as in section 4.1. The experiments were conducted with the reduction in the number of data for the minor classes of 90\% and the hyper parameters of our proposed method are described in Table 3. Fig. 9 depicts examples of transformed images xr when the minor classes are reduced by 90\%.

\section{Conclusions}
To eﬀectively use simulated images, for deep learning, it is required to accurately transform the simulated images to the real images. The state-of-the-art method, CycleGAN, would not be sensitive enough to the various change in pose and shape of subjects, especially when the training data are imbalanced. To overcome this insensitiveness problem , propose to actively introduce the label information into the training process of CycleGAN. The eﬀectiveness of our proposed method called Label-CycleGAN is demonstrated through experiments using imbalanced data for image-to-image transformation.

{\small
\bibliographystyle{ieee_fullname}
\bibliography{egbib}
}

\end{document}